%% file: main.tex
\documentclass[journal]{IEEEtran}
\usepackage{amsmath,amsfonts}
\usepackage{algorithmic}
\usepackage{algorithm}
\usepackage{array}
\usepackage[caption=false,font=normalsize,labelfont=sf,textfont=sf]{subfig}
\usepackage{textcomp}
\usepackage{stfloats}
\usepackage{url}
\usepackage{verbatim}
\usepackage{graphicx}
\usepackage{cite}
\usepackage{comment}
\usepackage{newtxtext}
\usepackage{newtxmath}
\usepackage{capt-of}
\usepackage{booktabs}
\begin{document}

\title{
Co-training with Ego-centric Video and Demonstration for Robot Navigation Task
}

\author{
    Shoya Kuno$^{1,2}$, Yumo Ouchi$^{2}$, and Kanata Suzuki$^{2}$
    \thanks{
        $^{1}$Shoya Kuno is affiliated with Depertment of Informatics, Graduate School of Informatics, Kyoto University, Sakyo, Kyoto, Japan. 
        $^{2}$All authors are affiliated with Spatial Robotics Research Center, Fujitsu Limited.,
        Kanagawa 211-8588, Japan.
    }
}



\maketitle


\input{sub/0_abstract.tex}

\input{sub/1_introduction.tex}

\input{sub/2_related_work.tex}

\input{sub/3_proposed_method.tex}

\input{sub/4_experiments.tex}

\input{sub/5_results.tex}

\input{sub/6_conclusion.tex}

\section*{Acknowledgments}
This work was supported by JST PRESTO, Japan, Grant Number JPMJPR24T4.

\bibliographystyle{IEEEtran}
\bibliography{refs}

\end{document}

%% file: sub/0_abstract.tex
\begin{abstract}

    Vision-language-action (VLA) models are promising for diverse robotic tasks, but their performance heavily depends on large-scale high-quality training data, whose collection on real robots is costly and time-consuming. While prior work has explored augmenting manipulation datasets with egocentric human videos, applying such approaches to mobile robot navigation remains challenging due to viewpoint changes during locomotion. In this paper, we propose a framework that converts egocentric walking videos into datasets for mobile robot imitation learning. The proposed method estimates camera motion from human videos and transforms it into action representations compatible with ground mobile robots. By jointly training a VLA model on human-derived and robot-collected datasets, the model achieves improved language understanding and more robust action generation than training with either data source alone. Experiments on a fruit-search navigation task demonstrate that human egocentric videos provide an effective and scalable data source for mobile robot learning.
\end{abstract}

%% file: sub/1_introduction.tex
\section{Introduction}

\IEEEPARstart{S}{ervice} robots are expected to operate in unstructured real-world environments such as households and public spaces.
Imitation learning~\cite{imitation_learning_servey, imitation_learning_servey2} enables robots to acquire complex behaviors directly from demonstrations, and recent Vision-Language-Action (VLA) models~\cite{openvla, pi0, rt1, rt2} have shown strong potential for language-conditioned robot control across diverse tasks.

However, VLA models require large-scale high-quality robot datasets~\cite{largescalemultirobotlearning, bridgeData}, whose collection is costly and time-consuming.
In contrast, egocentric human videos~\cite{ego4d} are abundantly available, motivating their use as an alternative data source.
While prior studies have explored such approaches for manipulation tasks, directly applying them to mobile robot navigation remains challenging due to unstable viewpoint changes during human locomotion.
Therefore, human trajectories must be transformed into motion representations compatible with mobile robot kinematics before being used for imitation learning.

To address this issue, we propose a framework that converts egocentric walking videos into datasets for mobile robot imitation learning.
The proposed method estimates camera motion from videos and transforms the resulting trajectories into robot-compatible action commands through kinematic projection and trajectory smoothing.
Furthermore, by combining human-derived data with robot-collected data, the proposed framework achieves both scalability and robot-specific action consistency.
Experiments on a real robot demonstrate that the proposed method enables robust language-conditioned navigation in unseen environments.

The main contributions of this work are as follows:
\begin{itemize}
    \item A framework for converting egocentric walking videos into mobile robot imitation learning datasets.
    \item A trajectory transformation method that generates robot-compatible action representations from human locomotion.
    \item Experimental validation showing improved language-conditioned navigation through joint training with human and robot data.
\end{itemize}

\begin{figure*}[!t]
  \centering
  \includegraphics[width=1.0 \textwidth]{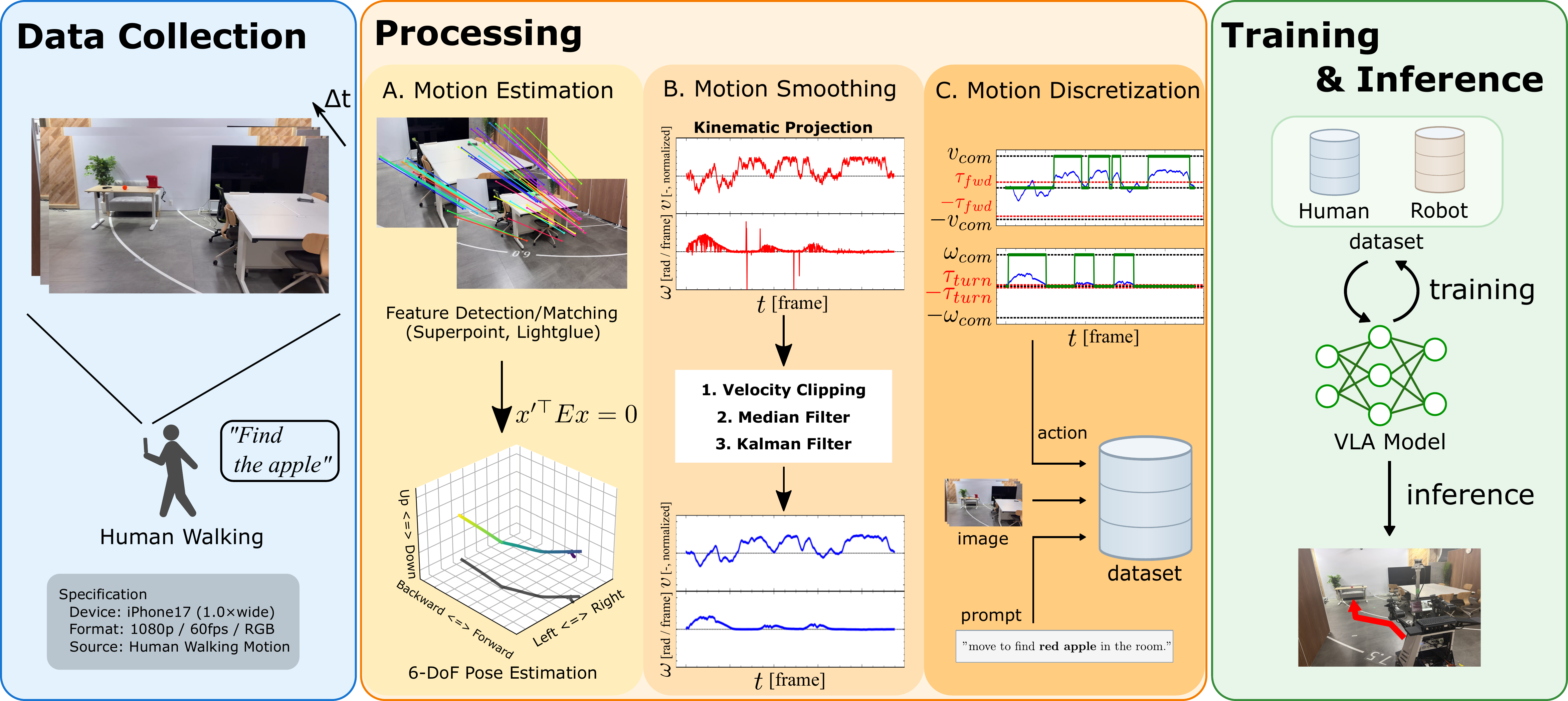}
    \caption{
        Overview of the proposed framework.
        Egocentric videos captured during human walking are processed to estimate camera motion, which is converted into robot-compatible action commands through trajectory decomposition and smoothing.
        The generated human-derived dataset is combined with robot-collected data to train a VLA model for language-conditioned mobile robot navigation.
    }
  \label{fig:system_overview}
\end{figure*}

%% file: sub/2_related_work.tex
\section{Related Work}
\subsection{Foundation-Model-based Navigation}
Recently, foundation-model-based navigation systems leveraging Vision-Language Models (VLMs) and Large Language Models (LLMs) have been actively studied, enabling semantic navigation and open-vocabulary object navigation~\cite{NavGPT}.
More recently, hierarchical navigation methods that separate semantic spatial representation from local trajectory generation have been proposed, integrating high-level subgoal generation using LLMs/VLMs with low-level trajectory planning and motion generation.
For example, LM-Nav~\cite{lmnav} decomposes language instructions into landmark-based subgoals using LLM, and realizes navigation by combining vision-language representations with a learned navigation policy.
In addition, VLMaps~\cite{vlmaps} constructs a spatial map embedded with vision-language features, enabling localization and navigation toward target objects specified in natural language.
These methods have demonstrated strong semantic understanding and generalization capabilities in unseen environments.
However, many existing approaches mainly focus on high-level semantic map construction and subgoal planning, while local trajectory generation and low-level control are often delegated to conventional navigation pipelines.
In contrast, our work aims to directly learn robotic navigation behaviors from large-scale egocentric human videos, enabling trajectory generation without relying on explicit subgoal planning.

\subsection{Imitation Learning from Human Data}
Imitation learning has been widely studied toward realizing robots capable of generating appropriate actions in unstructured environments.
Recently, action generation methods based on diffusion models and Vision-Language-Action (VLA) models, which jointly handle vision, language, and actions, have been proposed, enabling accurate and flexible action generation.
However, these approaches heavily rely on large-scale robotic datasets and multimodal web data, making scalable data collection a critical challenge.
In particular, collecting data using real robots is both costly and time-consuming, motivating approaches that leverage human behavioral data.
For example, Qian et al. constructed large-scale datasets by estimating action labels from videos using inverse kinematics and related techniques, although their approach requires complex model architectures during the estimation process.
In addition, Zheng et al. reported improved scalability and generalization performance by utilizing large-scale egocentric human videos.
Nevertheless, these methods learn representations in a human-centric latent space and do not directly model action representations suitable for robot control.
As a result, direct transfer to robotic systems remains difficult, particularly due to the domain gap caused by differences in viewpoints and motion characteristics between humans and robots.

To address the domain gap between humans and robots, our work converts egocentric walking videos into robot-compatible action labels using monocular visual odometry~\cite{monocular_visual_odometry}.

%% file: sub/3_proposed_method.tex
\section{Proposed Method}
This section presents a framework for generating robot learning datasets of image--action pairs from egocentric human videos.
The proposed framework consists of the following steps:

\begin{enumerate}
    \item \textbf{Visual Motion Estimation (Sec.~\ref{subsec:VO}):}
    A 6-DoF camera trajectory is estimated using SuperPoint and LightGlue.

    \item \textbf{Motion Smoothing and Kinematic Projection (Sec.~\ref{subsec:Motion-Smooth}):}
    The estimated trajectory is smoothed and projected onto the motion constraints of ground mobile robots.

    \item \textbf{Action Discretization (Sec.~\ref{subsec:Action-Discretization}):}
    The projected motion is converted into discrete robot actions.
\end{enumerate}

Finally, Sec.~\ref{subsec:VLA-Arch} describes a VLA-based imitation learning framework trained on the generated dataset.

\subsection{Visual Motion Estimation from Human Videos}
\label{subsec:VO}
We estimate camera motion from egocentric videos using a monocular visual odometry (VO)~\cite{monocular_visual_odometry} pipeline based on SuperPoint~\cite{superpoint} and LightGlue~\cite{lightglue}.
Egocentric videos often contain motion blur and abrupt camera motion caused by human walking, making robust feature matching essential.
The pipeline extracts visual features from consecutive frames, matches them across frames, and estimates the camera motion through three stages: feature extraction, feature matching, and motion estimation.

\subsubsection{Feature Extraction}
To extract reliable visual features from egocentric videos, we employ SuperPoint~\cite{superpoint}.
SuperPoint is a self-supervised feature detector that jointly performs keypoint detection and descriptor extraction from images. 
It is implemented as a fully convolutional network (FCN) with a shared encoder and two decoder branches.
Compared with traditional feature detectors such as SIFT~\cite{sift} and ORB~\cite{orb}, SuperPoint provides more robust keypoint detection and descriptor extraction under diverse environmental conditions.

\subsubsection{Feature Matching}
The extracted keypoints and descriptors are then matched across frames using LightGlue~\cite{lightglue}.
LightGlue is a feature matching algorithm that establishes correspondences between keypoints detected in two images based on their descriptors. 
It consists of multiple Transformer-based layers that incorporate two types of attention mechanisms: self-attention and cross-attention.
Self-attention models relationships among keypoints within the same image, whereas cross-attention captures relationships between keypoints across different images. 
Through these attention mechanisms, LightGlue enables robust feature matching based on both spatial information and descriptor similarity.

\subsubsection{Motion Estimation}
Using the matched feature correspondences, the essential matrix $E$ is estimated to recover the camera motion~\cite{essential_mat}.
The essential matrix is a $3 \times 3$ matrix that describes the geometric relationship between two images. 
Given a point $x$ in the first image and its corresponding point $x'$ in the second image, their relationship is expressed as $x'^T E x = 0$. 
Since the essential matrix encodes the relative rotation and translation of the camera, it enables estimation of the camera motion.
To improve robustness, the essential matrix is estimated using the RANSAC algorithm~\cite{ransac}, which removes outlier correspondences caused by dynamic objects or environmental noise.
The relative camera poses estimated between consecutive frames are accumulated to reconstruct the full camera trajectory. 
The resulting trajectory serves as the basis for generating robot-compatible motion commands in the subsequent stage.

\subsection{Motion Smoothing and Kinematic Projection} \label{subsec:Motion-Smooth}
Although visual odometry provides full 6-DoF camera motion, ground mobile robots typically operate under nonholonomic constraints and primarily execute planar motion consisting of forward/backward translation and yaw rotation.
Therefore, the estimated camera motion must be converted into robot-compatible planar motion commands. 
This process transforms noisy egocentric motion into smooth velocity signals while enforcing the kinematic constraints of ground robots.
The resulting motion signals are then used to generate robot navigation commands in the subsequent action discretization stage.

\subsubsection{Kinematic Projection}


Let the relative pose between consecutive frames be denoted as $\Delta T_k \in SE(3)$:
\begin{align}
    \Delta T_k = \begin{bmatrix}
        R_k & t_k \\
        0^{\top} & 1
    \end{bmatrix}.
\end{align}
Based on the nonholonomic model of ground robots, this transformation is projected onto $SE(2)$ to extract the forward velocity $v_k$ and angular velocity $\omega_k$.
The forward velocity is obtained by projecting the translation vector $t_k$ onto the robot's forward direction vector $n_{\mathrm{fwd}}$:
\begin{align}
    v_k = \frac{n_{\mathrm{fwd}}^{\top} t_k}{\Delta t}.
\end{align}
The angular velocity is computed from the planar component of the rotation matrix:
\begin{align}
    \omega_k = \frac{\mathrm{atan2}(R_k(2,1), R_k(1,1))}{\Delta t}.
\end{align}

This projection removes motion components irrelevant to ground robot locomotion, such as vertical oscillations and pitch rotations caused by human walking.
Because monocular visual odometry suffers from scale ambiguity, the projected translational component is not treated as a metric linear velocity. Instead, it is interpreted as a normalized forward-motion signal indicating the relative strength of motion intention.
The signal is then clipped to a predefined range to suppress outliers caused by unstable egocentric motion, and subsequently mapped to the nominal linear velocity of the target robot in the action discretization stage.

\subsubsection{Velocity Clipping}
Robot motion is subject to physical constraints, and abrupt velocity changes rarely occur in natural locomotion. 
Therefore, the estimated velocities are clipped within predefined limits to eliminate physically implausible motion between consecutive frames:
\begin{align}
    v_k &\leftarrow \mathrm{clip}(v_k, v_{\min}, v_{\max}). \\
    \omega_k &\leftarrow \mathrm{clip}(\omega_k, \omega_{\min}, \omega_{\max}).
\end{align}
The clipping limits are determined based on the maximum linear and angular velocities of the target robot platform.
In our implementation, the limits are set to $v_{\min} = -0.3, v_{\max} = 0.3, \omega_{\min} = -0.5,$ and $\omega_{\max} = 0.5$.

\subsubsection{Median Filtering}
To suppress impulsive outliers in the instantaneous velocity estimates, a median filter with a window size of $N = 5$ is applied.
A median filter introduces a phase delay of $(N-1)/2$ frames. 
This delay is comparable to or smaller than the update rate commonly used in autonomous navigation control loops (e.g., 30\,Hz) and therefore does not significantly affect control performance.
The median filter removes impulsive outliers, while the Kalman filter provides temporally consistent smoothing.

\subsubsection{Kalman Filtering}

To further smooth the trajectory, a Kalman filter based on a constant-velocity model is applied to combine noisy observations with a motion prior.
The state vector is defined as
\begin{align}
    x_k = \begin{bmatrix}
        v_k \\
        \omega_k
    \end{bmatrix}.
\end{align}
The system is modeled as a linear discrete-time process:
\begin{align}
    x_k &= x_{k-1} + w_{k-1}, \\
    z_k &= x_k + n_k,
\end{align}
where $w \sim \mathcal{N}(0, Q)$ denotes the process noise and $n \sim \mathcal{N}(0, R)$ denotes the observation noise.
In our implementation, the noise parameters are set to $Q = 0.005$ and $R = 0.5$ (i.e., $R/Q = 100$). 
These parameters are empirically tuned to favor temporal smoothness while maintaining responsiveness to motion changes. 
This configuration results in a relatively small Kalman gain, emphasizing temporal smoothness and the historical velocity trend.

Overall, the proposed pipeline converts noisy 6-DoF egocentric motion into smooth planar velocity commands suitable for ground robots.

\subsection{Action Discretization for Target Robot Consistency}
\label{subsec:Action-Discretization}



This process converts egocentric motion signals into robot-compatible discrete navigation actions.
The filtered velocity estimates are mapped to the discrete action space of the target robot. 
To suppress noise caused by micro-oscillations during human walking or temporary hesitation, we employ a nonlinear discretization scheme that extracts essential navigation intent.

First, the forward/backward velocity component $v$ is discretized. 
In monocular visual odometry, the estimated translation suffers from scale ambiguity, making it difficult to recover physically meaningful linear velocities. 
Therefore, the magnitude of the estimated translation is interpreted as the \emph{strength of motion intention} of the recorder. 
This value is then mapped to the nominal speed of the target robot (0.2\,m/s). 
To obtain discrete navigation actions compatible with the robot controller, the velocity is converted using threshold-based discretization.
Specifically, the forward velocity $v$ is discretized using the following rule:
\begin{align}
    v_d = \begin{cases}
        +v_{\text{nom}}, & v > \tau_{\text{fwd}}, \\
        -v_{\text{nom}}, & v < \tau_{\text{bwd}}, \\
        0, & \text{otherwise}.
    \end{cases}
\end{align}
where $\tau_{\text{fwd}} = 0.3$ and $\tau_{\text{bwd}} = -1.5$. 
Backward motion typically reflects stronger intentional actions; therefore, a larger threshold is used.

Next, the rotational velocity component $\omega$ is discretized.
Unlike translation, the rotational component estimated by monocular VO preserves geometrically consistent rotation angles. 
Therefore, the frame-wise rotation angle (rad/frame) can be used as a direct indicator of turning motion.
The rotational velocity is discretized using the following rule:
\begin{align}
    \omega_d = \begin{cases}
        +\omega_{\text{turn}}, & \omega > \tau_{\text{turn}}, \\
        -\omega_{\text{turn}}, & \omega < -\tau_{\text{turn}}, \\
        0, & \text{otherwise}.
    \end{cases}
\end{align}
where $\tau_{\text{turn}} = 0.003$.
This threshold suppresses small fluctuations caused by body sway during walking or minor visual disturbances, enabling the extraction of meaningful turning intentions.


As a final post-processing step, a mode filter is applied to remove short-duration action reversals in the discrete action sequence.
Specifically, action transitions lasting fewer than 12 frames (approximately 200\,ms) are replaced with the preceding action label. 
This step enforces temporal consistency in the action sequence.
Considering the physical response time of mobile robot actuators, this filtering removes high-frequency control commands that are infeasible for real robots. 
Consequently, the resulting action sequence captures stable and consistent motion intentions aligned with the visual observations.

Through this pipeline, egocentric walking videos are transformed into a high-quality robot learning dataset of image–action pairs compatible with the kinematic constraints of the target robot.

\subsection{Imitation Learning on VLA Architecture} \label{subsec:VLA-Arch}
\begin{figure}[!t]
    \centering
    \includegraphics[width=\linewidth]{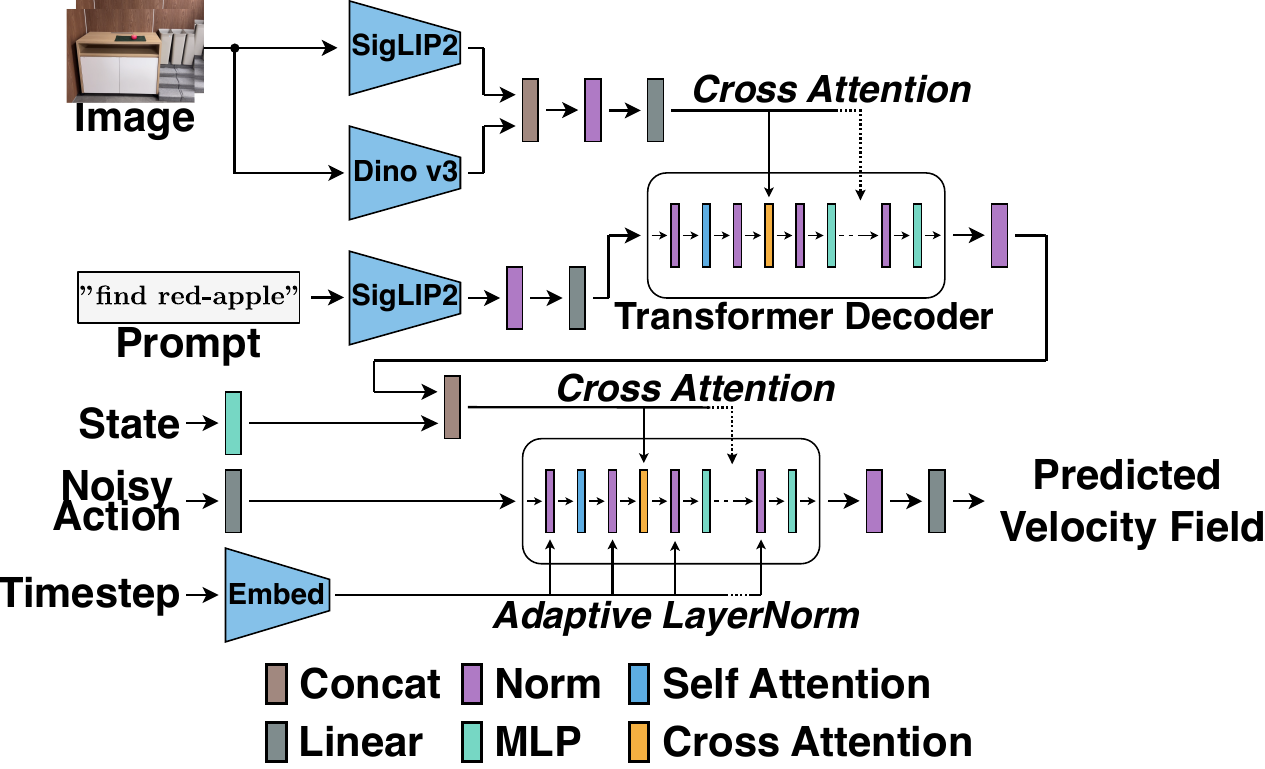}
    \caption{
    Overview of the VLA architecture. 
    A dual-encoder consisting of SigLIP2 and DINOv3 is employed to extract visual and language features, 
    which are integrated using a Transformer-based adapter. 
    The output tokens of the adapter are concatenated with state tokens and used as memory in the cross-attention layers of a Diffusion Transformer. 
    }
    \label{fig:model_arch}
\end{figure}
Using the dataset constructed in Sec.~\ref{subsec:VO}--\ref{subsec:Action-Discretization}, this section describes a Vision-Language-Action (VLA) model-based imitation learning framework that directly generates robot actions from visual and linguistic inputs.
The model takes as input a visual observation, a language instruction, the robot state, a noisy action sequence, and a diffusion time step.

We adopt the VLA architecture illustrated in Fig.~\ref{fig:model_arch}.
For the prompt encoder, we employ Siglip2~\cite{siglip2_vision} to encode natural language instructions into language tokens.
For visual encoding, we adopt a dual-encoder architecture consisting of Siglip2~\cite{siglip2_vision} and DINOv3~\cite{dinov3}. 
Siglip2 embeds images and text into a shared feature space, enabling aligned representations between visual inputs and language instructions. 
However, Siglip2 alone may not capture the fine-grained visual details required for robot action generation. 
To address this limitation, we incorporate DINOv3, which is trained in a self-supervised manner and provides high-resolution visual representations.
The visual encoders are kept frozen during training to preserve their pretrained representations and stabilize learning.
Image features are extracted from both Siglip2 and DINOv3, yielding two feature vectors that are fused via a linear combination.

The fused visual features and language features are fed into a Transformer-based adapter.
In this module, cross-attention is applied with the language features as queries and the visual features as keys and values, producing vision-conditioned language representations. 
These representations are then used as conditioning inputs for the downstream action generation module (action expert).



For action generation, we employ an 8-layer Transformer-based decoder, namely a Diffusion Transformer (DiT)~\cite{dit}. 
The model predicts a sequence of future actions over a fixed horizon of 32 steps.
We adopt conditional flow matching (CFM)~\cite{flow_matching} to deterministically generate actions from noise. 
Formally, the model learns a conditional velocity field $v_\theta(x_t, t \mid c)$, where $x_t$ denotes the noisy action, $t$ is the diffusion time step, and $c$ represents the conditioning inputs.
The DiT adopts a pre-normalization architecture, in which LayerNorm is applied before each sub-layer. 
This design improves training stability and mitigates vanishing gradients in deep Transformer networks.
Furthermore, the diffusion time-step embedding is incorporated into both the scaling and shifting parameters of Adaptive Layer Normalization (AdaLN), enabling time-dependent feature modulation.
The encoder features derived from visual and language inputs are injected into each layer of the DiT via cross-attention. 
This conditioning aligns the generated actions with both the visual observations and the natural language instructions.

Furthermore, to train the model, we construct a dataset by combining two types of data: 
(i) a human-derived dataset generated from egocentric walking videos, and 
(ii) a robot-collected dataset obtained from real-world robot executions.
The human-derived dataset enables scalable and low-cost data collection and provides diverse motion patterns, contributing to improved generalization of the VLA model. 
However, due to differences in camera properties such as field of view and viewpoint between egocentric videos and robot-mounted cameras, a domain gap exists between the two data sources.
To mitigate this gap, the robot-collected dataset is incorporated to expose the model to robot-specific visual observations and dynamics. 
By jointly training on both datasets, the model learns both generalizable motion representations and robot-grounded action patterns. 
This complementary design enables the model to leverage large-scale pseudo-demonstrations while maintaining consistency with real robot behavior.

Overall, the proposed VLA-based imitation learning framework enables direct action generation from visual and language inputs while effectively leveraging heterogeneous data sources to achieve both scalability and real-world applicability.
This unified framework bridges the gap between video-derived supervision and real-world robot execution, enabling scalable and practical policy learning.

%% file: sub/4_experiments.tex
\section{Experiments}
To comprehensively evaluate the proposed method, we conduct experiments on a fruit-search navigation task using both human-derived and robot-collected datasets.
All experiments are performed on \textit{CobotMagic}~\cite{cobotmagic}, a dual-arm mobile manipulator.
We evaluate the method from three perspectives: 
(i) stability of action generation, 
(ii) generalization to language prompts, and 
(iii) robustness to unseen initial positions.

\subsection{Task Setup}
As shown in Fig.~\ref{fig:task_overview}, we conduct a fruit-search navigation task in an indoor laboratory environment. 
The objective is for the robot to locate a specified fruit and navigate to a position directly in front of the target object.
The target objects consist of three types of fruits: a red apple, a green apple, and an orange, as illustrated in Fig.~\ref{fig:task_object}. 
Each object is placed at one of the predefined goal locations, $G_A$, $G_B$, or $G_C$.

\subsection{Datasets}
\label{subsec:dataset}
\subsubsection{Robot-collected dataset}
The robot-collected dataset was acquired in an indoor laboratory environment using CobotMagic.
The camera mounted on CobotMagic was fixed to the robot body.
The onboard camera recorded videos at 480×640 resolution and 30 fps. 
The robot executed the task at a linear velocity of 0.2\,m/s and an angular velocity of 0.4\,rad/s.
A total of 150 trials were conducted, with 50 trials collected for each goal location.
The target object for the navigation task was limited to a red apple.
The initial position of the task was randomly selected from four predefined locations: $S_A$, $S_B$, $S_C$, and $S_D$.

\subsubsection{Human-derived dataset}
The human-derived dataset is constructed from videos captured while a human performs the same task in the same environment.
The videos are recorded using an iPhone (standard camera application) at a resolution of 1080p and a frame rate of 60\,fps. 
The recordings include all three types of fruits (red apple, green apple, and orange) and are collected under two recording conditions, with 80 trials in total. 
The start and goal positions are uniformly distributed to avoid bias.
The recording conditions are as follows:
(i) A simple setting in which only the target fruit is present, with no distractor objects (60 trials).  
(ii) A more complex setting in which non-target fruits are also present, requiring the agent to ignore irrelevant objects (20 trials).
In total, 240 videos are collected and converted into image–action pair datasets using the method described in Sec.~\ref{subsec:VO}--\ref{subsec:Action-Discretization}.

\begin{figure}[!t]
    \centering
    \subfloat[Bird’s-eye view of the experimental environment.]{
        \includegraphics[width=0.85\linewidth]{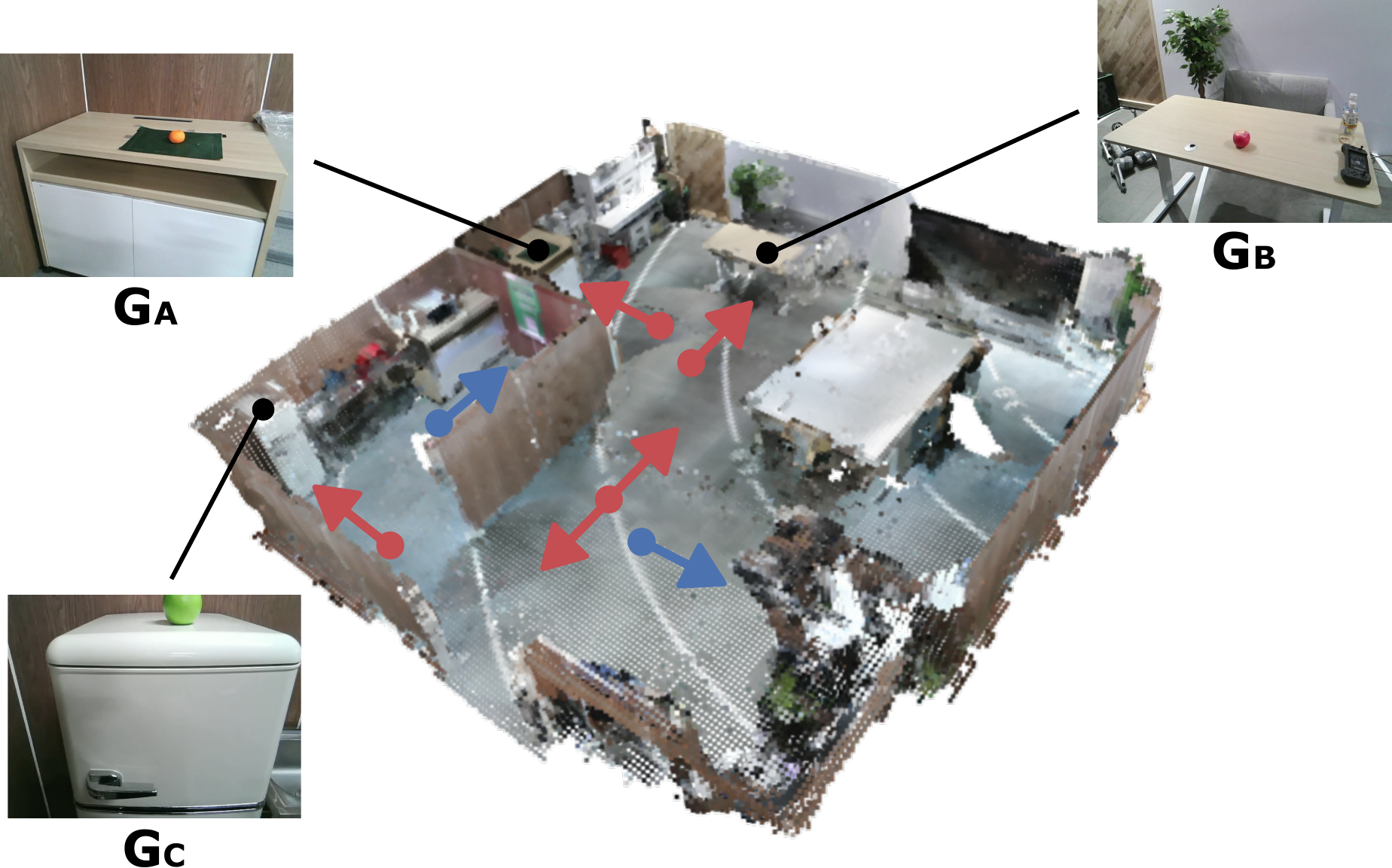}
    }
    \vspace{10pt}
    \subfloat[Abstract task configuration]{
        \includegraphics[width=0.6\linewidth]{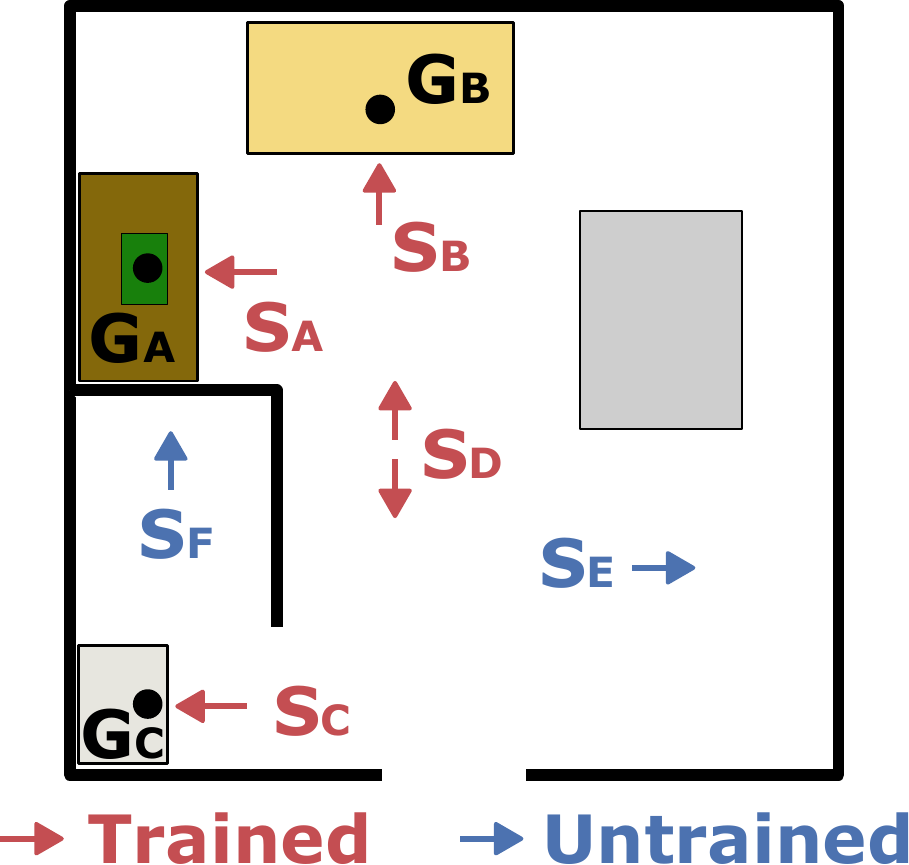}
    }    
    \caption{
        Overview of the experimental environment.
        (a) Bird’s-eye view of the reconstructed 3D environment.
        (b) Abstract task configuration.
        $S_A, S_B, S_C, S_D$ denote initial robot positions included in the training dataset,
        while $S_E, S_F$ denote unseen initial positions used for evaluation.
        In each trial, a target object (fruit) is placed at one of the goal locations
        $G_A, G_B, G_C$.
        Arrows indicate the initial robot headings.
    }
    \label{fig:task_overview}
    
    \vspace{10pt}
    
    \includegraphics[width=0.75\linewidth]{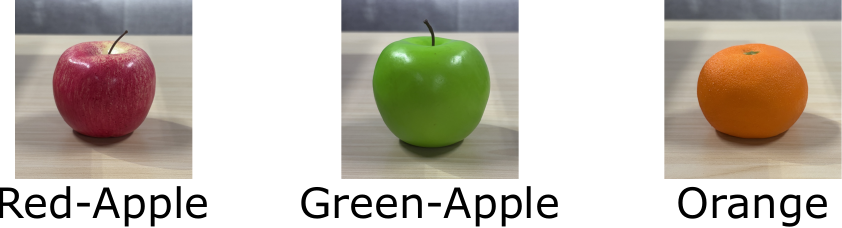}
    \caption{
    Target objects. 
Three types of food samples are used: a red apple, a green apple, and an orange.
    }
    \label{fig:task_object}
\end{figure}

\subsection{Training}
The proposed VLA model is trained via imitation learning on the dataset described in Sec.~\ref{subsec:dataset}. 
The overall architecture follows Fig.~\ref{fig:model_arch}.
For visual and language encoding, we use pretrained models, namely SigLIP2 (siglip2\_base\_256) and DINOv3 (vit\_base\_patch16\_dinov3), respectively. 
These encoders are kept frozen during training.
The Transformer-based adapter for fusing visual and language features consists of 8 layers, with a hidden dimension of 512 and a feedforward dimension of 2048. 
The action generation module, implemented as a Diffusion Transformer (DiT), adopts the same configuration.
We use AdamW with a learning rate of $1\times10^{-4}$ and cosine scheduling. 
Training is performed with a batch size of 4 and gradient accumulation over 32 steps for 400,000 iterations. 
Standard geometric and photometric augmentations are applied during training.
To prevent overfitting, data augmentation is applied to the input images. 
Geometric transformations include random rotation (±3°), translation (up to 5\% of the image size), and scaling (0.9--1.1). 
Photometric transformations include brightness and contrast variations (±30\%) and saturation adjustment (±10\%).

\subsection{Evaluation}
\subsubsection{Evaluation Metrics}

We compare three training configurations:
(i) robot-only training,
(ii) human-only training, and
(iii) joint training with both datasets.
A trial is considered successful if the robot reaches within 1\,m of the target object and remains stationary for more than 1\,s.
Collisions and stationary or rotational behaviors lasting longer than 5\,s are regarded as failures.
Task success rate is used as the evaluation metric.

\subsubsection{Evaluation from the trained initial position}
The robot is initialized from training positions $(S_A, S_B, S_C, S_D)$ and tasked with navigating toward a target object placed at one of $(G_A, G_B, G_C)$.
For $S_D$, the initial orientation is conditionally set according to the target goal, as shown in Fig.~\ref{fig:task_overview}.
This position is included to evaluate challenging transitions between constrained regions.
The target objects consist of a red apple, a green apple, and an orange.
We evaluate two settings:
(i) a simple setting containing only the target object, and
(ii) a cluttered setting containing additional non-target objects.
For each combination of initial and goal positions, two trials are performed.

\subsubsection{Evaluation from the untrained initial position}
To evaluate generalization, the robot is initialized from unseen positions $(S_E, S_F)$ that are not included in the training dataset.
The target object is placed at one of $(G_A, G_B, G_C)$, and only the red apple is used without distractor objects.
For each combination of initial and goal positions, two trials are performed.

%% file: sub/5_results.tex
\section{Results and Discussion}
\subsection{Evaluation from the trained initial position}
\begin{figure}[t]
    \centering
    \includegraphics[width=\linewidth]{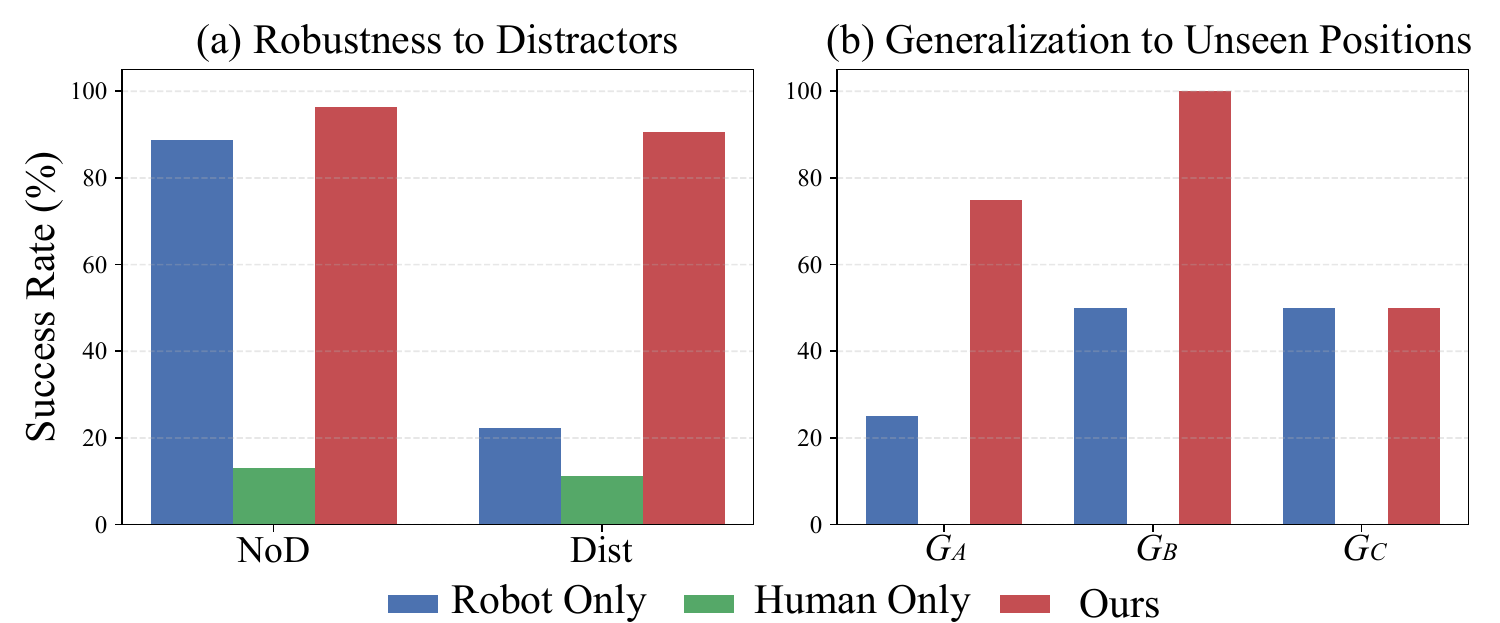}
    \caption{
        Average success rates for each method under different conditions.
        (a) Success rate (\%) under environments with and without distractors.
        (b) Success rate (\%) on unseen initial positions across different goal locations.
    }
    \label{fig:success_rate}
\end{figure}

Fig.~\ref{fig:success_rate}(a) shows the task success rates under two conditions: 
a setting without distractors (NoD), where only the target object is present, and a setting with distractors (Dist), where non-target objects are placed at other goal locations. 
The blue, green, and red bars represent models trained on the robot-collected dataset only, the human-derived dataset only, and the combined dataset, respectively. 
Detailed success rates for each object are reported in Table~\ref{tab:main_res}.

\begin{table}[t]
    \centering
    \caption{Success rate (\%) for each target object under different conditions.}
    \label{tab:main_res}
    \begin{tabular}{lcccccc}
        \toprule
         & \multicolumn{2}{c}{Red apple} & \multicolumn{2}{c}{Green apple} & \multicolumn{2}{c}{Orange} \\
        \cmidrule(lr){2-3} \cmidrule(lr){4-5} \cmidrule(lr){6-7}
        Method & NoD & Dist & NoD & Dist & NoD & Dist \\
        \midrule
        Robot Only & 88.9 & 22.2 & 88.9 & 22.2 & 88.9 & 22.2 \\
        Human Only & 11.1 & 11.1 & 22.2 & 11.1 & 5.56 & 11.1\\
        Ours & \textbf{100.0} & \textbf{88.9} & \textbf{88.9} & \textbf{88.9} & \textbf{100.0} & \textbf{94.4} \\
        \bottomrule
        \addlinespace
        \multicolumn{7}{l}{\small ※ NoD: No Distractors, Dist: With Distractor objects}
    \end{tabular}
\end{table}

In the NoD setting, the model trained only on the human-derived dataset achieves a low success rate of 11.1\%. In contrast, the robot-only model achieves 88.9\%, and the combined model achieves comparable or better performance. This result suggests that training solely on human-derived data leads to unstable action generation due to domain gaps in camera viewpoints and motion patterns.
For example, when initialized at $S_D$, the human-only model can reach $G_B$, which requires only forward motion. However, when the target is $G_A$, the model fails to generate the required sequence of left rotation followed by forward motion and instead continues rotating near $G_B$. Similarly, when the target is $G_C$, the model fails to generate right rotation and cannot move toward the target. 
In contrast, the combined model achieves higher success rates. This improvement can be attributed to the integration of diverse viewpoints and motion patterns, leading to more generalizable representations. 
Additionally, the combined model exhibits fewer collisions near $S_C$, indicating improved stability.

In the Dist setting, the success rate decreases for both the human-only and robot-only models. 
In particular, the robot-only model often misidentifies distractors as targets and fails to generate behaviors that avoid them.
The success rate does not drop to 0\% in some cases because, for trajectories such as $S_D \rightarrow G_B$ or $G_C$, other goal locations do not appear within the robot's field of view.
Even for the combined model, performance degrades compared to the NoD setting. 
This degradation is most pronounced when initialized at $S_C$, where the target occupies only a small region in the image, making recognition difficult. 
As shown in Fig.~\ref{fig:task_overview}, the small object size leads to confusion between visually similar objects, such as red and green apples.
In contrast, when the target is an orange, the model generates appropriate actions, indicating that the language instruction is correctly understood.
These results demonstrate that incorporating human-derived data improves the model’s ability to generate actions conditioned on language instructions.

\begin{figure*}[t]
    \centering
    \includegraphics[width=0.95\linewidth]{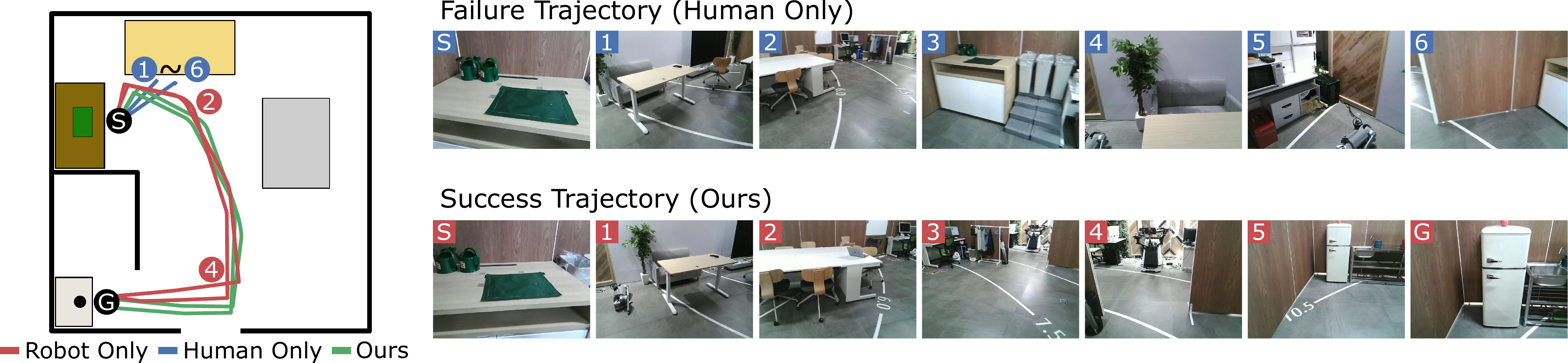}
    \caption{
        Trajectory comparison and front-camera observation transitions.
        The left figure shows the robot trajectories, where the numbered markers correspond to the front-camera observations shown on the right.
    }
    \label{fig:trajectory_image}
\end{figure*}

Finally, Fig.~\ref{fig:trajectory_image} shows the robot trajectories and transitions of front-camera observations for the exploration task from $S_A$ to $G_C$.
As described above, the model trained only on human-derived data fails the task by continuing to rotate near $G_B$ (Fig.~\ref{fig:trajectory_image}, 1$\sim$6).
In contrast, both the robot-only model and the combined model generate smooth trajectories.
These results indicate that the proposed method improves performance by incorporating human-derived data without degrading the smoothness of action generation.

\begin{table}[t]
    \centering
    \caption{Success rate (\%) when using initial positions ($S_E$, $S_F$) not included in the training dataset. The input language is ``Move to find a red apple in the room.''}
    \label{tab:untrain_pos}
    \begin{tabular}{lccccccc}
        \toprule
        & \multicolumn{3}{c}{$S_E$} & \multicolumn{3}{c}{$S_F$} & Total \\
        \cmidrule(lr){2-4} \cmidrule(lr){5-7}
        Method & $G_A$ & $G_B$ & $G_C$ & $G_A$ & $G_B$ & $G_C$ &  \\
        \midrule
        Robot Only & 1/2 & 2/2 & 2/2 & 0/2 & 0/2 & 0/2 & 41.7\% \\
        Human Only & 0/2 & 0/2 & 0/2 & 0/2 & 0/2 & 0/2 & 0.0\% \\
        Ours       & \textbf{2/2} & \textbf{2/2} & \textbf{2/2} & \textbf{1/2} & \textbf{2/2} & \textbf{0/2} & \textbf{75.0\%}\\
        \bottomrule
    \end{tabular}
\end{table}




\subsection{Evaluation from the untrained initial position}
Fig.~\ref{fig:success_rate}(b) shows the task success rates for unseen initial positions. 
Detailed success rates for each goal location are reported in Table~\ref{tab:untrain_pos}.
The model trained only on the human-derived dataset achieves a success rate of 0\% for all combinations of initial positions and goal locations, indicating that training solely on human-derived data is insufficient for generalization.
When initialized at $S_E$, both the robot-only and combined models achieve high success rates. 
However, the robot-only model exhibits one failure due to a collision with a table located near $S_E$.
In contrast, when initialized at $S_F$, the robot-only model fails in all cases, resulting in a success rate of 0\%, whereas the combined model achieves 50\%. 
The robot-only model fails to identify a feasible forward direction and remains rotating in place. 
In contrast, the combined model successfully explores and identifies feasible directions, enabling task completion in some cases.
These results demonstrate that combining robot-collected and human-derived data improves both the stability of action generation and exploration capability under unseen initial conditions.

\subsection{Discussion}
These results indicate that robot-collected and human-derived data play distinct and complementary roles in model training. 
Human-derived data provides diverse motion patterns and flexible responses to language instructions, improving generalization to unseen environments. 
In contrast, robot-collected data captures robot-specific observation and motion characteristics, such as camera viewpoints and actuation dynamics, thereby improving the stability of action generation.
However, using only human-derived data is insufficient for direct robot control due to domain gaps in camera viewpoints and motion characteristics. 
The proposed approach mitigates this gap by incorporating robot-collected data, enabling the model to retain the benefits of human-derived data while adapting to real-world robotic environments. 
Furthermore, human-derived data can be collected at significantly lower cost and with greater ease than robot-collected data, enabling scalable dataset construction.
These findings demonstrate that effective robot learning datasets can be constructed from human egocentric videos, and that combining such data with robot-collected data enables both scalability and real-world applicability.

However, several limitations remain.
First, the proposed method requires robot-collected data to mitigate the domain gap, and its performance may degrade when such data are insufficient or unavailable.
Second, the optimal mixing ratio between human-derived and robot-collected data has not been thoroughly investigated, and the amount of robot data required remains unclear.
Third, the quality of the generated dataset depends on the stability and coverage of egocentric videos, which can vary significantly depending on the recorder’s motion and recording conditions.
Furthermore, since this study focuses on a 2-DoF ground mobile robot, the gap between human motion and robot kinematics is only partially addressed, leaving room for improvement in action consistency.
Nevertheless, the core idea of the proposed method is to estimate motion from egocentric videos using Visual Odometry and Essential Matrix decomposition.
Therefore, by converting the estimated motion according to the kinematic model of each platform, the proposed framework can potentially be extended to other robotic platforms, such as omnidirectional mobile robots and humanoid robots.

%% file: sub/6_conclusion.tex
\section{Conclusion}
In this paper, we propose a framework that converts egocentric videos captured during human walking into datasets suitable for imitation learning using monocular visual odometry.
Furthermore, we demonstrate that combining the resulting human-derived dataset with robot-collected data enables the development of a VLA model that achieves both scalability and real-world applicability.
Experimental results show that the model trained on the combined dataset improves both generalization to unseen environments and the stability of action generation, outperforming models trained on either dataset alone.
These findings indicate that human egocentric videos are an effective and scalable data source for training mobile robots.

%% file: refs.bib
@inproceedings{superpoint,
author = {DeTone, Daniel and Malisiewicz, Tomasz and Rabinovich, Andrew},
title = {SuperPoint: Self-Supervised Interest Point Detection and Description},
booktitle = {Proceedings of the IEEE Conference on Computer Vision and Pattern Recognition (CVPR) Workshops},
month = {June},
year = {2018}
}

@inproceedings{lightglue,
author = {Lindenberger, Philipp and Sarlin, Paul-Edouard and Pollefeys, Marc},
title = {LightGlue: Local Feature Matching at Light Speed},
booktitle = {Proceedings of the IEEE/CVF International Conference on Computer Vision (ICCV)},
month = {October},
year = {2023},
pages = {17627-17638}
}

@inproceedings{siglip2_vision,
author = {Zhai, Xiaohua and Mustafa, Basil and Kolesnikov, Alexander and Beyer, Lucas},
title = {Sigmoid Loss for Language Image Pre-Training},
booktitle = {Proceedings of the IEEE/CVF International Conference on Computer Vision (ICCV)},
month = {October},
year = {2023},
pages = {11975-11986}
}

@misc{dinov3,title         = {DINOv3},author        = {Oriane Siméoni and Huy V. Vo and Maximilian Seitzer and Federico Baldassarre and Maxime Oquab and Cijo Jose and Vasil Khalidov and Marc Szafraniec and Seungeun Yi and Michaël Ramamonjisoa and Francisco Massa and Daniel Haziza and Luca Wehrstedt and Jianyuan Wang and Timothée Darcet and Théo Moutakanni and Leonel Sentana and Claire Roberts and Andrea Vedaldi and Jamie Tolan and John Brandt and Camille Couprie and Julien Mairal and Hervé Jégou and Patrick Labatut and Piotr Bojanowski},year          = {2025},eprint        = {2508.10104},archiveprefix = {arXiv},primaryclass  = {cs.CV},url           = {https://arxiv.org/abs/2508.10104}}

@misc{flow_matching,title         = {Flow Matching for Generative Modeling},author        = {Yaron Lipman and Ricky T. Q. Chen and Heli Ben-Hamu and Maximilian Nickel and Matt Le},year          = {2023},eprint        = {2210.02747},archiveprefix = {arXiv},primaryclass  = {cs.LG},url           = {https://arxiv.org/abs/2210.02747}}

@misc{pi0,title={$\pi_0$: A Vision-Language-Action Flow Model for General Robot Control},author={Kevin Black and Noah Brown and Danny Driess and Adnan Esmail and Michael Equi and Chelsea Finn and Niccolo Fusai and Lachy Groom and Karol Hausman and Brian Ichter and Szymon Jakubczak and Tim Jones and Liyiming Ke and Sergey Levine and Adrian Li-Bell and Mohith Mothukuri and Suraj Nair and Karl Pertsch and Lucy Xiaoyang Shi and James Tanner and Quan Vuong and Anna Walling and Haohuan Wang and Ury Zhilinsky},year={2026},eprint={2410.24164},archivePrefix={arXiv},primaryClass={cs.LG},url={https://arxiv.org/abs/2410.24164},}

@misc{rt1,title={RT-1: Robotics Transformer for Real-World Control at Scale},author={Anthony Brohan and Noah Brown and Justice Carbajal and Yevgen Chebotar and Joseph Dabis and Chelsea Finn and Keerthana Gopalakrishnan and Karol Hausman and Alex Herzog and Jasmine Hsu and Julian Ibarz and Brian Ichter and Alex Irpan and Tomas Jackson and Sally Jesmonth and Nikhil J Joshi and Ryan Julian and Dmitry Kalashnikov and Yuheng Kuang and Isabel Leal and Kuang-Huei Lee and Sergey Levine and Yao Lu and Utsav Malla and Deeksha Manjunath and Igor Mordatch and Ofir Nachum and Carolina Parada and Jodilyn Peralta and Emily Perez and Karl Pertsch and Jornell Quiambao and Kanishka Rao and Michael Ryoo and Grecia Salazar and Pannag Sanketi and Kevin Sayed and Jaspiar Singh and Sumedh Sontakke and Austin Stone and Clayton Tan and Huong Tran and Vincent Vanhoucke and Steve Vega and Quan Vuong and Fei Xia and Ted Xiao and Peng Xu and Sichun Xu and Tianhe Yu and Brianna Zitkovich},year={2023},eprint={2212.06817},archivePrefix={arXiv},primaryClass={cs.RO},url={https://arxiv.org/abs/2212.06817},}

@inproceedings{rt2,title={RT-2: Vision-Language-Action Models Transfer Web Knowledge to Robotic Control},author={Zitkovich, Brianna and et al.},booktitle={CoRL},year={2023}}

@misc{openvla,title={OpenVLA: An Open-Source Vision-Language-Action Model},author={Moo Jin Kim and Karl Pertsch and Siddharth Karamcheti and Ted Xiao and Ashwin Balakrishna and Suraj Nair and Rafael Rafailov and Ethan Foster and Grace Lam and Pannag Sanketi and Quan Vuong and Thomas Kollar and Benjamin Burchfiel and Russ Tedrake and Dorsa Sadigh and Sergey Levine and Percy Liang and Chelsea Finn},year={2024},eprint={2406.09246},archivePrefix={arXiv},primaryClass={cs.RO},url={https://arxiv.org/abs/2406.09246},}

@ARTICLE{monocular_visual_odometry,author={Françani, André O. and Maximo, Marcos R. O. A.},journal={IEEE Access},title={Transformer-Based Model for Monocular Visual Odometry: A Video Understanding Approach},year={2025},volume={13},number={},pages={13959-13971},doi={10.1109/ACCESS.2025.3531667}}

@article{imitation_learning_servey,author  = {Fang, Bin and Jia, Shidong and Guo, Di and Xu, Muhua and Wen, Shuhuan and Sun, Fuchun},title   = {Survey of imitation learning for robotic manipulation},journal = {International Journal of Intelligent Robotics and Applications},year    = {2019},volume  = {3},number  = {4},pages   = {362--369},doi     = {10.1007/s41315-019-00103-5},url     = {https://doi.org/10.1007/s41315-019-00103-5},issn    = {2366-598X},}

@ARTICLE{imitation_learning_servey2,author={Zare, Maryam and Kebria, Parham M. and Khosravi, Abbas and Nahavandi, Saeid},journal={IEEE Transactions on Cybernetics},title={A Survey of Imitation Learning: Algorithms, Recent Developments, and Challenges},year={2024},volume={54},number={12},pages={7173-7186},doi={10.1109/TCYB.2024.3395626}}

@misc{largescalemultirobotlearning,title={RoboNet: Large-Scale Multi-Robot Learning},author={Sudeep Dasari and Frederik Ebert and Stephen Tian and Suraj Nair and Bernadette Bucher and Karl Schmeckpeper and Siddharth Singh and Sergey Levine and Chelsea Finn},year={2020},eprint={1910.11215},archivePrefix={arXiv},primaryClass={cs.RO},url={https://arxiv.org/abs/1910.11215},}

@InProceedings{bridgeData,title = 	 {BridgeData V2: A Dataset for Robot Learning at Scale},author =       {Walke, Homer Rich and Black, Kevin and Zhao, Tony Z. and Vuong, Quan and Zheng, Chongyi and Hansen-Estruch, Philippe and He, Andre Wang and Myers, Vivek and Kim, Moo Jin and Du, Max and Lee, Abraham and Fang, Kuan and Finn, Chelsea and Levine, Sergey},booktitle = 	 {Proceedings of The 7th Conference on Robot Learning},pages = 	 {1723--1736},year = 	 {2023},editor = 	 {Tan, Jie and Toussaint, Marc and Darvish, Kourosh},volume = 	 {229},series = 	 {Proceedings of Machine Learning Research},month = 	 {06--09 Nov},publisher =    {PMLR},url = 	 {https://proceedings.mlr.press/v229/walke23a.html},}

@InProceedings{ego4d,author={Grauman, Kristen and et al.},title     = {Ego4D: Around the World in 3,000 Hours of Egocentric Video},booktitle = {Proceedings of the IEEE/CVF Conference on Computer Vision and Pattern Recognition (CVPR)},month     = {June},year      = {2022},pages     = {18995-19012}}

@misc{sift,author = {Lindeberg, Tony},institution = {KTH, Computational Biology, CB},note = {QC 20120524},number = {5},pages = {10491--},title = {Scale invariant feature transform},volume = {7},DOI = {10.4249/scholarpedia.10491},URL = {http://www.scholarpedia.org/article/Scale_Invariant_Feature_Transform},year = {2012}}

@INPROCEEDINGS{orb,author={Rublee, Ethan and Rabaud, Vincent and Konolige, Kurt and Bradski, Gary},booktitle={2011 International Conference on Computer Vision},title={ORB: An efficient alternative to SIFT or SURF},year={2011},volume={},number={},pages={2564-2571},doi={10.1109/ICCV.2011.6126544}}

@article{ransac,author = {Fischler, Martin A. and Bolles, Robert C.},title = {Random sample consensus: a paradigm for model fitting with applications to image analysis and automated cartography},year = {1981},issue_date = {June 1981},publisher = {Association for Computing Machinery},address = {New York, NY, USA},volume = {24},number = {6},issn = {0001-0782},url = {https://doi.org/10.1145/358669.358692},doi = {10.1145/358669.358692},journal = {Commun. ACM},month = jun,pages = {381–395},numpages = {15}}

@book{essential_mat,title={Multiple View Geometry in Computer Vision},author={Hartley, Richard and Zisserman, Andrew},year={2003},publisher={Cambridge University Press},edition={Second},isbn={0521540518}}

@misc{dit,title={Scalable Diffusion Models with Transformers},author={William Peebles and Saining Xie},year={2023},eprint={2212.09748},archivePrefix={arXiv},primaryClass={cs.CV},url={https://arxiv.org/abs/2212.09748},}

@INPROCEEDINGS{vlmaps,
  author={Huang, Chenguang and Mees, Oier and Zeng, Andy and Burgard, Wolfram},
  booktitle={2023 IEEE International Conference on Robotics and Automation (ICRA)}, 
  title={Visual Language Maps for Robot Navigation}, 
  year={2023},
  volume={},
  number={},
  pages={10608-10615},
  doi={10.1109/ICRA48891.2023.10160969}}

@InProceedings{lmnav,
  title = 	 {LM-Nav: Robotic Navigation with Large Pre-Trained Models of Language, Vision, and Action},
  author =       {Shah, Dhruv and Osi\'nski, B\l{a}\.zej and ichter, brian and Levine, Sergey},
  booktitle = 	 {Proceedings of The 6th Conference on Robot Learning},
  pages = 	 {492--504},
  year = 	 {2023},
  editor = 	 {Liu, Karen and Kulic, Dana and Ichnowski, Jeff},
  volume = 	 {205},
  series = 	 {Proceedings of Machine Learning Research},
  month = 	 {14--18 Dec},
  publisher =    {PMLR},
  url = 	 {https://proceedings.mlr.press/v205/shah23b.html}
}

@misc{cobotmagic,
    author={AgileX},
    title={COBOT Magic},
    note={\url{https://global.agilex.ai/products/cobot-magic}},
    year={Accessed:2026-02-21}
    }

@article{NavGPT, title={NavGPT: Explicit Reasoning in Vision-and-Language Navigation with Large Language Models}, volume={38}, url={https://ojs.aaai.org/index.php/AAAI/article/view/28597}, DOI={10.1609/aaai.v38i7.28597}, abstractNote={Trained with an unprecedented scale of data, large language models (LLMs) like ChatGPT and GPT-4 exhibit the emergence of significant reasoning abilities from model scaling. Such a trend underscored the potential of training LLMs with unlimited language data, advancing the development of a universal embodied agent. In this work, we introduce the NavGPT, a purely LLM-based instruction-following navigation agent, to reveal the reasoning capability of GPT models in complex embodied scenes by performing zero-shot sequential action prediction for vision-and-language navigation (VLN). At each step, NavGPT takes the textual descriptions of visual observations, navigation history, and future explorable directions as inputs to reason the agent’s current status, and makes the decision to approach the target. Through comprehensive experiments, we demonstrate NavGPT can explicitly perform high-level planning for navigation, including decomposing instruction into sub-goals, integrating commonsense knowledge relevant to navigation task resolution, identifying landmarks from observed scenes, tracking navigation progress, and adapting to exceptions with plan adjustment. Furthermore, we show that LLMs is capable of generating high-quality navigational instructions from observations and actions along a path, as well as drawing accurate top-down metric trajectory given the agent’s navigation history. Despite the performance of using NavGPT to zero-shot R2R tasks still falling short of trained models, we suggest adapting multi-modality inputs for LLMs to use as visual navigation agents and applying the explicit reasoning of LLMs to benefit learning-based models. Code is available at: https://github.com/GengzeZhou/NavGPT.}, number={7}, journal={Proceedings of the AAAI Conference on Artificial Intelligence}, author={Zhou, Gengze and Hong, Yicong and Wu, Qi}, year={2024}, month={Mar.}, pages={7641–7649} }
